\documentclass[letterpaper]{article} 
\usepackage{booktabs}
\usepackage{multirow}

\usepackage{aaai24}  
\usepackage{times}  
\usepackage{helvet}  
\usepackage{courier}  
\usepackage[hyphens]{url}  
\usepackage{graphicx} 
\usepackage{natbib}  
\usepackage{caption} 
\usepackage{algorithm}
\usepackage{algorithmic}

\usepackage{newfloat}
\usepackage{listings}
\DeclareCaptionStyle{ruled}{labelfont=normalfont,labelsep=colon,strut=off} 
\floatstyle{ruled}
\newfloat{listing}{tb}{lst}{}
\floatname{listing}{Listing}
\title{All Stories Are One Story: Emotional Arc Guided Procedural Game Level Generation
}

\author {
    Yunge Wen\equalcontrib, Chenliang Huang\equalcontrib, Hangyu Zhou, Zhuo Zeng, Chun Ming Louis Po, Julian Togelius, Timothy Merino, Sam Earle
}

\affiliations {
    New York University – Tandon School of Engineering\\
    Brooklyn, New York\\
    yw3776@nyu.edu, ch4955@nyu.edu, hz3735@nyu.edu, zz5295@nyu.edu, cmp743@nyu.edu, julian@togelius.com, tm3477@nyu.edu, sam.earle@nyu.edu
}

\makeatletter
\def\@copyright{}
\def\@copyrightspace{}
\makeatother

\begin{document}

\maketitle
\maketitle

\begin{abstract}

The emotional arc is a universal narrative structure underlying stories across cultures and media—an idea central to structuralist narratology, often encapsulated in the phrase “all stories are one story.” We present a framework for procedural game narrative generation that incorporates emotional arcs as a structural backbone for both story progression and gameplay dynamics. Leveraging established narratological theories and large-scale empirical analyses, we focus on two core emotional patterns—\textit{Rise} and \textit{Fall}—to guide the generation of branching story graphs. Each story node is automatically populated with characters, items, and gameplay-relevant attributes (e.g., health, attack), with difficulty adjusted according to the emotional trajectory. Implemented in a prototype action role-playing game (ARPG), our system demonstrates how emotional arcs can be operationalized using large language models (LLMs) and adaptive entity generation. Evaluation through player ratings, interviews, and sentiment analysis shows that emotional arc integration significantly enhances engagement, narrative coherence, and emotional impact. These results highlight the potential of emotionally structured procedural generation for advancing interactive storytelling for games.

\end{abstract}

\section{Introduction}

Narrative structure plays a critical role in shaping human experience across media. In storytelling traditions ranging from ancient myths to modern cinema, emotional arcs—patterns of rise and fall in affective tone—have emerged as a universal framework for organizing plot progression and audience engagement. An empirical study \cite{reagan2016emotional} validated these structures at scale, showing that emotional arcs such as \textit{Cinderella} (Rise-Fall-Rise) or \textit{Rags to Riches} (Rise) recur across genres and cultures. While such arcs have long influenced hand-authored narratives in games, their integration into procedural content generation (PCG) remains limited.

As large language models (LLMs) gain prominence in game AI, enabling dynamic generation of storylines, dialogue, and even levels, questions arise about how to ensure that generated narratives exhibit global coherence and emotional depth. However, most existing work focuses on local narrative fluency or branching dialogue, often without attention to overarching emotional structure. 

In this paper, we propose a framework that explicitly integrates emotional arcs into procedural story and level generation. We model narrative progression as a directed acyclic graph (DAG), with each node corresponding to a distinct emotional state—either \textit{Rise} or \textit{Fall}—and automatically populated with story content, characters, gameplay attributes, and difficulty parameters. Our system is implemented within a prototype action role-playing game (ARPG), where the generated narrative graph is translated into a sequence of interconnected dungeons reflecting the intended emotional trajectory.

We address two core research questions:
\begin{itemize}
\item \textbf{RQ1:} Does embedding emotional arcs into procedural game generation enhance player experience?
\item \textbf{RQ2:} Are emotional arcs perceptible to players and computational emotion models in procedurally generated content?
\end{itemize}

Through a mixed-method evaluation involving user ratings, qualitative interviews, and automated sentiment analysis, we demonstrate that emotional arc integration improves enjoyment, supports emotionally coherent gameplay, and is both perceptible and impactful. These findings suggest that emotional structure can serve as a generative constraint that enhances both narrative quality and user experience in AI-driven game design.

\section{Related Works}

\subsection{Emotional Arcs: The Building Blocks of All Stories}

The idea that “all stories are one story” has long been explored in narratology, particularly within structuralist and transmedial frameworks. \cite{propp2010morphology, campbell2008hero, polti1921thirtysix} This concept refers to the existence of a universal narrative structure that underlies diverse stories across cultures and media, emphasizing shared human experiences, archetypal roles, and recurring themes. 

A foundational work in this area is Vladimir Propp’s \textit{Morphology of the Folktale} (1928), in which he analyzed over 100 Russian folktales and identified a consistent set of narrative functions that appear in a fixed sequence. These include elements such as departure, initiation, and return, forming the structural backbone of many traditional narratives. Building on this, Joseph Campbell introduced the concept of the Monomyth, or Hero’s Journey, in \textit{The Hero with a Thousand Faces} (1949). His model outlines a cyclical path where the hero leaves the ordinary world, undergoes trials, achieves transformation, and ultimately returns. This structure has significantly influenced modern storytelling, particularly in literature and film. 

In more recent work, researchers have applied statistical methods to narrative theory. Reagan et al. (2016) used machine learning techniques to analyze a large corpus of English fiction, focusing on emotional arcs, which represent the sentiment trajectories experienced by readers as a story progresses. By applying singular value decomposition, hierarchical clustering, and self-organizing maps, they identified six fundamental emotional arc types. These arcs represent combinations of \textit{Rise} and \textit{Fall} emotional patterns and serve as foundational components in a wide variety of narratives, ranging from \textit{Cinderella} to \textit{The Magic of Oz}.

The evolution of video games from rule-based, abstract mechanics to richly mimetic representations reflects a broader shift toward narrativization. Unlike linear storytelling in film, which often relies on structures such as the Hero’s Journey, the commercial success of games like \textit{Until Dawn} \cite{untildawn2015} and \textit{The Last of Us} \cite{lastofus2013} suggests a strong audience appetite for branching narrative structures, which support emotional engagement and replayability. These effects are achieved through the flexible and dynamic use of narrative arcs that adapt to player choice.

Our narrative arc generation framework directly builds upon the Rise and Fall concepts introduced in Reagan et al. (2016), using these emotional trajectories to inform our narrative generation prompting pipeline and guide the design of interactive, emotionally resonant game experiences.

\subsection{Procedural Narrative \& Large Language Models}

Narrative graphs are widely used in game storytelling to represent possible player story paths as directed graphs, where nodes denote narrative units (e.g., events, scenes) and edges encode temporal, causal, or choice-based relations. \cite{ware2022multiagent,riedl2013interactive,riedl2010narrative, liapis2013designer} They support dynamic traversal via planning or search algorithms, enabling integration of story with gameplay and player interaction. \cite{alvarez2022story, grabska2021application} While human-authored game narratives require significant resources and expertise, PCG aims to ease this burden.

Recent advances in LLMs have shifted narrative generation toward data-driven paradigms. \cite{ammanabrolu2020story, buongiorno2024pangea, bubeck2023sparks, calderwood2022spinning} LLMs exhibit strong semantic coherence, world knowledge, and dialogue capabilities, making them suitable for dynamic, adaptive story generation. For instance, GENEVA \cite{leandro2024geneva} uses GPT-4o-mini to generate narrative beats that form a directed acyclic graph, enabling controllable emotional progression and branching storylines. NarrativeGenie \cite{kumaran2024narrativegenie} augments this by integrating real-time LLM-driven character dialogue that responds to player choices, aligning narrative dynamics with gameplay.

LLMs have also been applied to procedural level generation. Prior work demonstrates that models like GPT-2 and GPT-3 can generate tile-based levels in games such as \textit{Super Mario Bros.} and \textit{Sokoban} by learning design patterns from structured data \cite{sudhakaran2023mariogpt, LevelGenerationLLM}. Word2World \cite{nasir2024word2world} further shows how narrative prompts can be grounded into level design by mapping story semantics to spatial tile layouts and NPC placements.

Our approach builds upon these developments by using high-level emotional arcs to control global narrative direction, while synchronizing with adaptive gameplay difficulty. This enables a unified structure where the evolving narrative directly modulates player experience, providing both emotional immersion and challenge pacing.

\begin{figure*}[t]
\centering
\includegraphics[width=\textwidth]{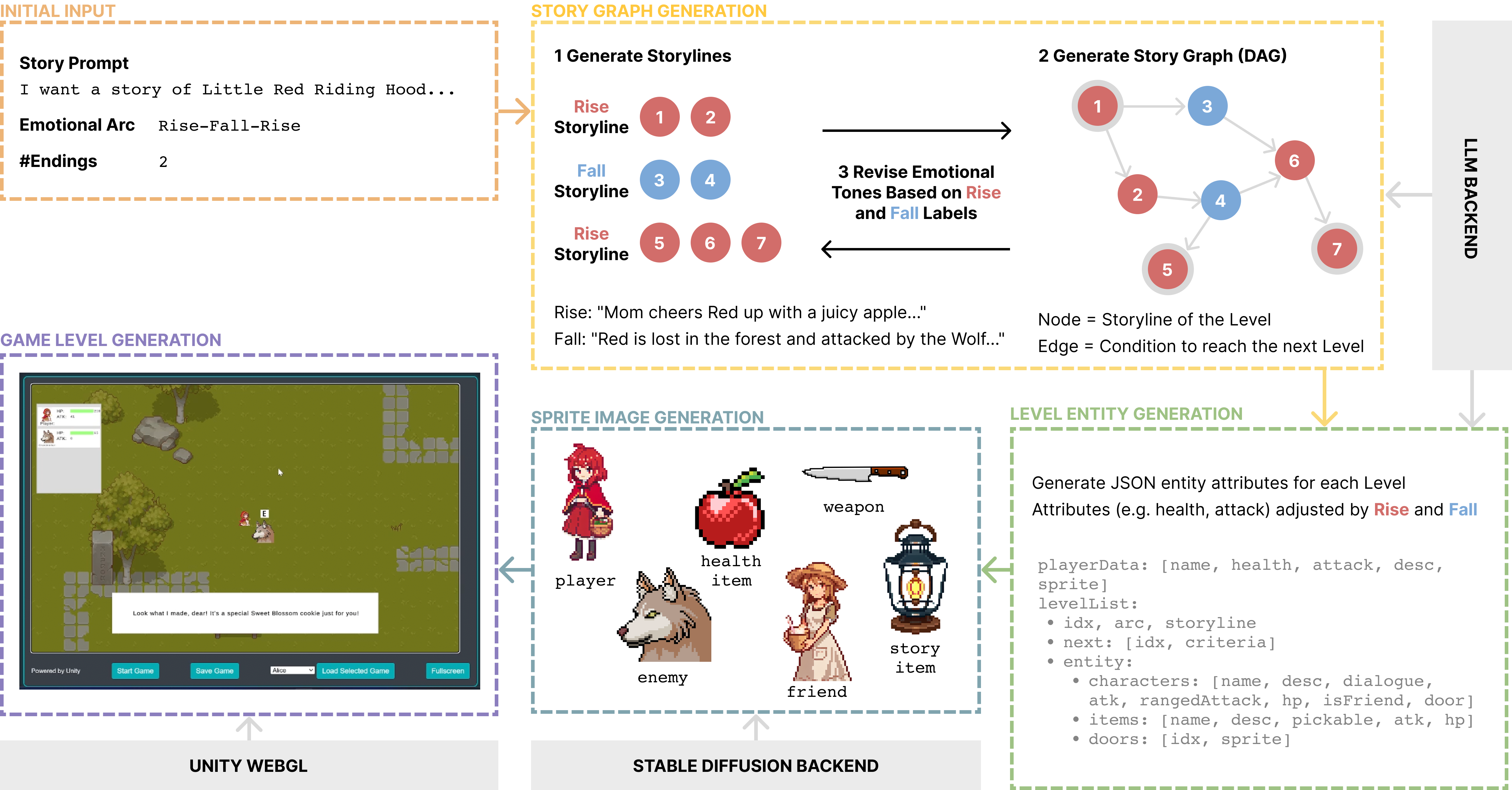}
\caption{
End-to-end framework for emotional arc–guided game generation. The pipeline takes a user-provided narrative prompt and emotional arc as input, constructs a branching story graph with labeled emotional phases (Rise, Fall), and generates corresponding game levels by instantiating structured entities, sprite assets, and gameplay logic. The system integrates a language model backend for narrative generation and a diffusion model for visual assets, and outputs playable games via Unity WebGL.}
\label{fig:pipeline}
\end{figure*}

\section{Method}
Building upon prior work in narrative generation and emotional arc modeling, we present a pipeline designed to automatically generate interconnected game levels from high-level user input, grounded in emotionally structured story arcs. The complete framework, as illustrated in Figure \ref{fig:pipeline}, is designed to:

\begin{itemize}
    \item Allow users to input a concise narrative prompt and emotional arc type, and generate a branching story graph labeled with emotional progression.
    \item Automatically convert each story node into a corresponding game level, extracting relevant characters and items, along with their semantic descriptions and gameplay-relevant attributes (e.g., attack, health).
    \item Provide an interactive interface that enables designers to revise both the storyline and the entity configurations during the generation process, allowing for human-in-the-loop customization before final game level construction.
\end{itemize}
The output of the pipeline is a structured JSON file for each node in the story graph, containing storyline descriptions, emotional arc labels, player goals, characters (NPCs, enemies), items, and gameplay attributes. These outputs can be further adapted to various gameplay templates.

\subsection{Story Generation}

Our framework begins with a narrative generation module that transforms a high-level user-provided story prompt into a branching story graph shaped by a predefined emotional arc.

We adopt GPT-4o-mini as the core language model for generation due to its superior instruction-following ability, contextual understanding, and low latency. To address the challenge of overwhelming the LLM with complex, multi-step generation tasks, we employ an AI Chaining \cite{WuChains2022} strategy. This decomposes the generation process into a sequence of focused subtasks, with intermediate validations to enable greater control over narrative structure.

To model emotional trajectory in narrative structure, we utilize the framework proposed by Reagan et al. (2016), which classifies stories into six fundamental arcs: \textit{Rags to Riches} (Rise), \textit{Tragedy} (Fall), \textit{Man in a Hole} (Fall → Rise), \textit{Icarus} (Rise → Fall), \textit{Cinderella} (Rise → Fall → Rise), and \textit{Oedipus }(Fall → Rise → Fall). Its compositional structure based on Rise and Fall patterns makes it particularly suited for dynamic content generation and difficulty pacing. Compared to other structuralist models in narratology, this emotional arc framework offers higher flexibility and adaptability across genres and gameplay styles.

At the beginning of the pipeline, the user is asked to input a short narrative prompt of no more than 30 words, select one of the six emotional arcs, and optionally specify the minimum number of desired story endings (defaulting to one). 

Based on this input, the Story Graph Generation module produces a logically interconnected branching structure in the form of a directed acyclic graph. The resulting graph consists of nodes and edges:

\begin{itemize}
    \item Story Nodes: Each node is assigned a structural label---either Rise or Fall---to indicate its intended narrative trajectory within the overall arc. At this stage, the focus is on defining the node's role and position in the story progression rather than specifying detailed emotional text.
    \item Story Edges: Edges between nodes represent the criteria for progressing from one story beat to another. These are described in natural language and grounded in simple player interactions, such as “talk to a friendly NPC,” “pick up an item,” or “defeat an enemy.” Regardless of whether a path is linear or branching, each transition is defined by a clear, actionable trigger that can be mapped onto in-game mechanics.
\end{itemize}


Once the story graph is generated, it is visualized through a D3.js-based web interface. Users can revise the emotional label or storyline of each node, modify transition criteria, add or remove nodes, and rewire the graph structure.


Upon finalization, the modified story graph is passed back to the system, which performs semantic refinement to inject the emotional essence of the rise or fall arc into each node’s textual content. This phase transforms the structural labels into concrete narrative expressions, ensuring that the storyline tone aligns with the intended emotional trajectory:

\begin{quote}
\scriptsize
\ttfamily
mind\_reset(Rise) = "Embrace an uplifting and hopeful tone, highlighting progress and positive transformation." \\
mind\_reset(Fall) = "Adopt a more somber and challenging tone, emphasizing setbacks and internal or external conflicts." \\

StoryChainRevision( \\
\hspace*{1em}mind\_reset,    // Set emotional tone \\
\hspace*{1em}story\_list,    // List of previous story nodes \\
\hspace*{1em}current\_story\_node    // Current node to polish \\
)
\end{quote}

This bi-directional interaction between user control and model-guided validation not only ensures structural coherence but also imbues each storyline with authentic emotional depth that embodies the intended arc.

\subsection{Entity Generation}

Once the story graph is finalized, each story node is passed to the Level Entity Generation phase, which identifies and instantiates the game-relevant entities in a modular and extensible format. This design enables flexible customization by game designers and supports seamless integration into a variety of game genres. The generated structure conforms to the following schema:

\begin{quote}
\scriptsize
\ttfamily
playerData: \{name, health, attack, desc, sprite\} \\
levelList: \\
\hspace*{2em}- idx, arc, storyline \\
\hspace*{2em}- next: \{idx, criteria\} \\
\hspace*{2em}- entity: \\
\hspace*{4em}- NPCs: \{name, desc, dialogue[], atk, ranged, hp, friend, door\} \\
\hspace*{4em}- items: \{name, desc, pickable, atk, hp\} \\
\hspace*{4em}- doors: \{idx, sprite\}
\end{quote}

Designers may freely extend these templates with additional properties, based on design requirements or genre-specific needs. This modular representation facilitates efficient mapping from narrative content to gameplay mechanics.

A central design objective of the framework is to align gameplay difficulty with the underlying emotional trajectory of the story. Nodes associated with a Rise arc are designed to be more optimistic and approachable, whereas Fall arcs introduce heightened tension and challenge. To operationalize this alignment, entity statistics (e.g., \texttt{health}, \texttt{attack}) are automatically configured based on the emotional arc and the narrative progression stage. Designers retain full control to override or extend these automatically generated values, allowing for a balance between automated narrative-aware tuning and manual creative input. This dual-mode mechanism supports both dynamic difficulty balancing and designer-driven customization. A case study demonstrating this adaptive generation within an action role-playing game context is presented in the Case Study section.

To maintain narrative consistency and gameplay continuity, the Level Entity Generation phase processes story nodes in a sequential, level-wise fashion. For each node, the entity generation module considers not only the current narrative context but also the full history of prior nodes and generated entities. This context-aware mechanism ensures logical coherence across levels. For instance, if a character acquires a sword in one level, subsequent levels will reflect this state change by updating the character’s description and attributes. In the absence of new changes, entity states persist unaltered. This design preserves narrative causality while ensuring smooth gameplay transitions, tightly coupling story progression with game logic.

\begin{figure*}[t]
\centering
\includegraphics[width=\linewidth]{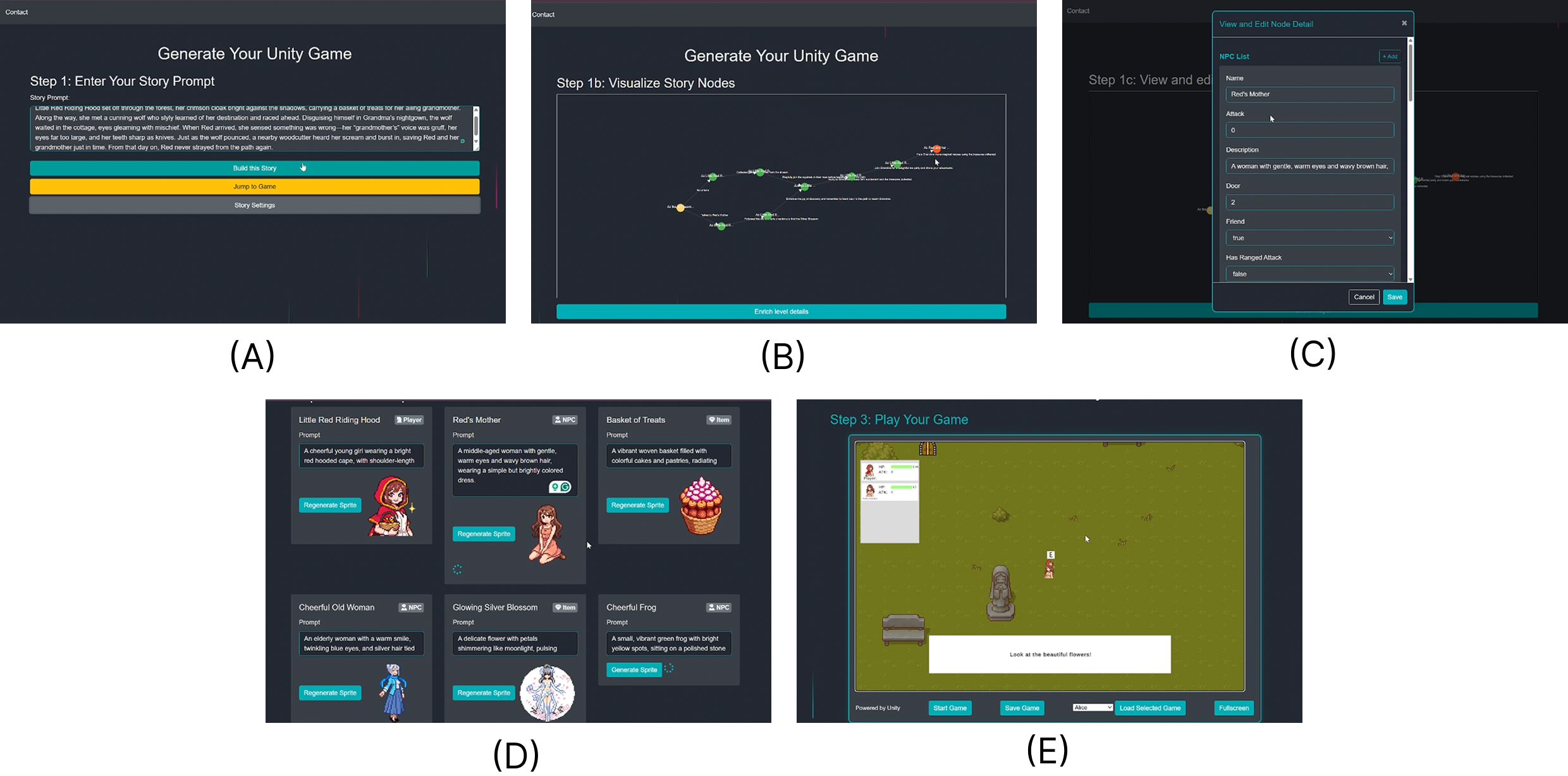}  
  \caption{End-to-end narrative generation and gameplay interface. (A) Users input a story prompt to generate an emotional arc. (B) A visualized story graph is created with nodes as narrative beats. (C) Each node’s content including NPCs and attributes like attack and health can be edited. (D) Pixel-art sprites are generated via a diffusion model with support for regeneration. (E) The resulting Unity game maps each node to a dungeon room aligned with the emotional trajectory.}
\label{fig:game}
\end{figure*}

\section{Case Study}

To evaluate the practical applicability of our emotional arc–driven narrative framework, we implemented a prototype action RPG (Figure \ref{fig:game}). The ARPG genre was chosen for its modular structure and ease of content decomposition. Its blend of narrative progression and dynamic difficulty makes it a suitable testbed for emotional arc-guided procedural generation.

In our implementation, each node in the generated story graph corresponds to a self-contained dungeon room. The game entities associated with that node—such as NPCs, enemies, and props—are placed within the room based on the story description. Navigation to subsequent nodes is implemented through doors, which may be gated by specific in-game interactions (e.g., completing a conversation or collecting an item). This design allows for both branching progression and conditional access, directly reflecting the structure of the underlying story graph.

The entire generation process is automated, beginning from the emotional arc-guided story graph. Each node is serialized into a structured JSON file, which includes its emotional label (Rise or Fall), storyline, player objectives, and a list of game entities with associated gameplay-relevant attributes such as \verb|attack|, \verb|health|, and \verb|hasRangedAttack|. These values, produced within the narrative generation phase, allow the large language model to modulate in-game entity strength according to the emotional tone and narrative phase.

To support visual representation, we integrate LayerDiffusion \cite{zhang2024transparentimagelayerdiffusion} with a pre-trained Stable Diffusion v1.5 to generate pixel-art style sprites for game entities. The model produces assets with transparent backgrounds that can be directly imported into the Unity engine. Users can interact with the generation interface to regenerate sprites if the default style does not match their preferences, enhancing personalization and replayability.

In Unity, these entities are instantiated using a set of reusable C\# base classes—\verb|Player|, \verb|NPC|, \verb|Item|, and \verb|Dialogue|—that are populated directly from the JSON. The architecture supports modular extension, allowing these components to carry over or evolve across different story nodes while maintaining narrative consistency.

To validate how emotional arcs influence gameplay dynamics, we analyzed difficulty as expressed through game entities. Enemies in Fall nodes were more likely to have higher \verb|attack|, \verb|health|, and ranged capabilities, while Rise nodes featured lower difficulty, friendlier NPCs, and less hostile environments. These patterns emerged solely from prompt engineering, demonstrating the model’s ability to align narrative pacing with gameplay challenge.

\section{Evaluation}

A mixed-method evaluation combining quantitative and qualitative techniques was conducted to assess the impact of emotional arc–guided game level generation. The evaluation was designed to address the following research questions:

\begin{itemize}
    \item \textbf{RQ1}: Does embedding emotional arcs into procedural
game generation enhance player experience?
    \item \textbf{RQ2}: Are emotional arcs perceptible to players and
computational emotion models in procedurally generated
content?
\end{itemize}
To investigate these questions, an open-ended exploratory user study was conducted, comparing gameplay experiences with and without emotional arc–driven narratives. RQ1 was examined through user-reported scales and post-task interviews, while RQ2 was assessed via binary identification tasks and sentiment analysis.

\subsection{User Study}

Sixteen participants (P1–P16) were recruited through email outreach and snowball sampling. The average age was 25.38 years (SD = 3.54; range = 21–34). Participants had academic backgrounds in Computer Science, Finance, Education, Marketing, among others. Prior gaming experience was self-reported using a four-level categorical scale: \textit{Never}, \textit{Rarely}, \textit{Often}, and \textit{Experienced}. Informed consent was obtained, and all data collection adhered to ethical data protection protocols.

Each session lasted approximately 45 minutes. After a tutorial on the system interface and gameplay mechanics, participants provided a narrative prompt, which was used to generate two distinct game episodes:

\begin{itemize}
    \item \textbf{Baseline}: A graph with coherent story progression, with all emotional arc labels set to None. No post-generation revision was performed to adjust the node storylines, resulting in a flat, affect-neutral narrative without rise or fall dynamics.

    \item \textbf{Emotional Arc}: One of the six canonical emotional arcs was randomly assigned. The generation process followed the procedure described in the Method section.
\end{itemize}

The order of these conditions was counterbalanced to mitigate order effects. Participants were not informed of the condition in each session. Both episodes were constrained to seven story nodes, with each node containing a maximum of 50 words. A single ending was enforced to reduce cognitive load. Sprite generation, gameplay mechanics, and difficulty parameters were held constant across conditions to isolate the narrative structure as the independent variable.

After completing both episodes, participants filled out a post-task Likert-scale questionnaire and indicated which session they believed contained an emotional arc. This was followed by a semi-structured interview to gather qualitative feedback on perceived differences and system quality.

Multiple metrics were employed to assess the effect of emotional arc in game level generation, including subjective user ratings, arc perceptibility, and objective sentiment-based validation.

\paragraph{User-Rated Scales.}
To address RQ1, a 7-point Likert scale was used to measure (1) \textit{Enjoyment}: How enjoyable was the game experience? (2) \textit{Relevance}: How relevant was the game to the initial story prompt? (3) \textit{Difficulty}: How challenging was the game?

Mean scores and 95\% confidence intervals were computed via bootstrapping with 1,000 samples. 

A one-sided Wilcoxon signed-rank test was used under the directional hypothesis: Emotional Arc \( > \) Baseline. Effect sizes were computed as \( r = \frac{|z|}{\sqrt{n}} \), where \( z \) is the z-score approximation.

\paragraph{Arc Perception.}
To address RQ2, participants identified which session they believed included an emotional arc (binary response: \textit{Session 1} / \textit{Session 2}).

\paragraph{Semi-Structured Interviews.}
Participants reflected on differences between conditions, narrative quality, and overall impressions of the gameplay experience.

\subsection{System-Level Analysis} 
To further address RQ2, a post-hoc sentiment analysis was conducted to evaluate alignment between intended emotional arcs and generated narratives. For each of the six arc types, 10 story graphs were generated using the same prompt set. Each graph contained exactly 7 nodes and was linearized by level index.

The GoEmotions multi-label classifier was used to predict 27 emotion categories for each node. Labels with probabilities $\geq 0.1$ were retained and mapped to valence scores (+1 = positive, 0 = neutral, -1 = negative). A composite valence score for each node was computed as $ V = \sum_i p_i \cdot v_i $, where $ p_i $ is the predicted probability and $ v_i \in \{-1, 0, +1\} $ is the mapped valence score. The resulting valence trajectory was plotted by level index. For each arc type, the average of 10 trajectories was computed to examine overall shape consistency.

\begin{figure}[t]
\centering
\includegraphics[width=\columnwidth]{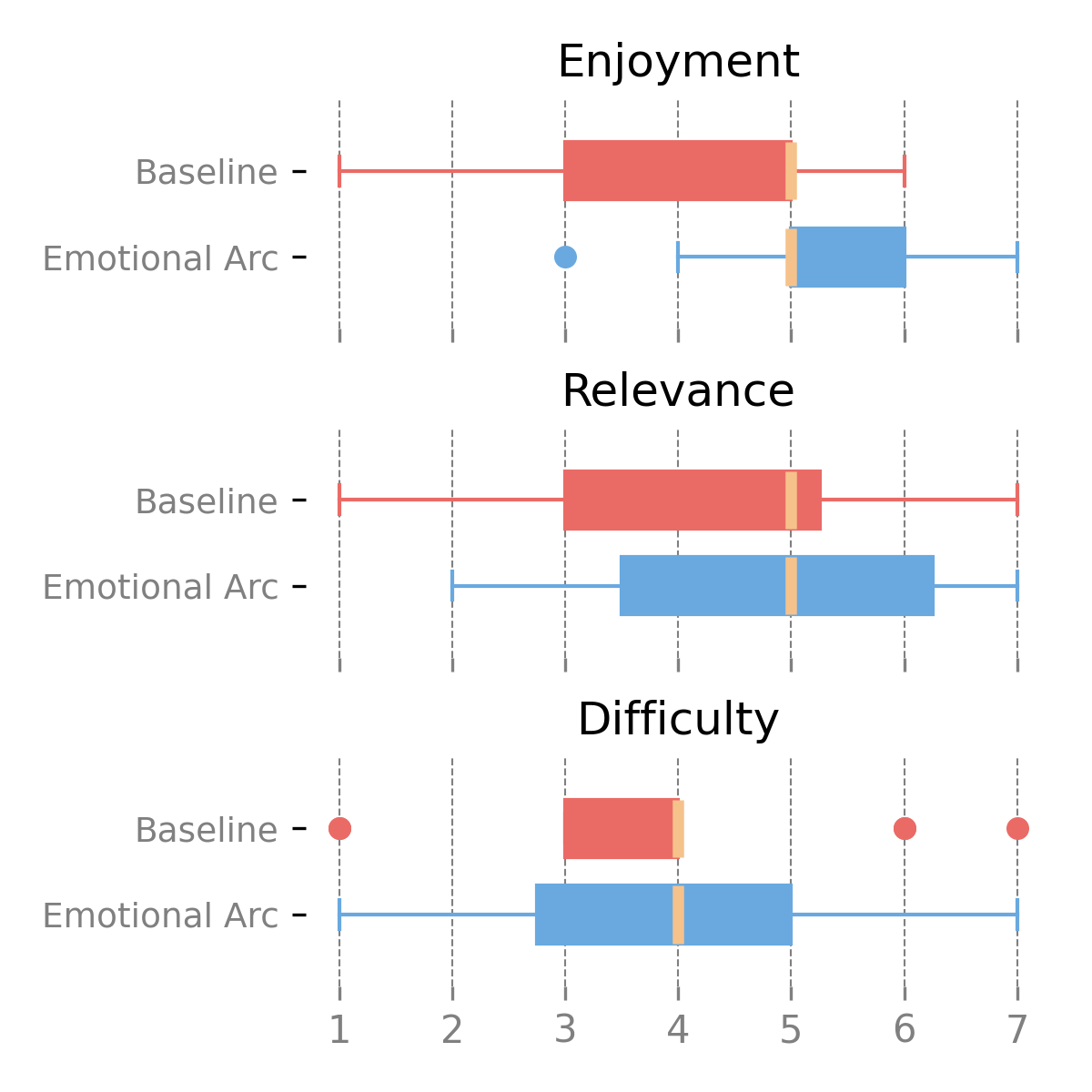}  
  \caption{Boxplot comparison of user ratings across three dimensions: \textit{Enjoyment}, \textit{Relevance}, and \textit{Difficulty}, under Baseline (red) and Emotional Arc (blue) conditions. Each box represents the interquartile range (IQR) with orange lines indicating medians. Colored dots denote mean values, and vertical dashed lines mark Likert scale levels (1–7). Emotional Arc shows a higher and more concentrated distribution for Enjoyment, while Relevance and Difficulty exhibit greater variance across conditions.}
\label{fig:user_boxplot}
\end{figure}

\begin{table}[t]
\centering
\small
\begin{tabular}{l l c c}
\toprule
\textbf{Metric} & \textbf{Condition} & \textbf{Mean} & \textbf{95\% CI} \\
\midrule
Enjoyment& Baseline      & 4.265 & [3.500, 4.938] \\
    & Emotional Arc & 5.261 & [4.812, 5.750] \\
\midrule
Relevance& Baseline      & 4.450 & [3.625, 5.188] \\
    & Emotional Arc & 4.749 & [3.750, 5.625] \\
\midrule
Difficulty& Baseline      & 3.794 & [3.000, 4.500] \\
    & Emotional Arc & 3.949 & [3.188, 4.689] \\
\bottomrule
\end{tabular}
\caption{Mean user ratings and 95\% confidence intervals (bootstrapped with 10,000 samples) for each metric under Baseline and Emotional Arc conditions.}
\label{tab:bootstrap_fancy}
\end{table}

\begin{table}[t]
\centering
\begin{tabular}{lccc}
\toprule
\textbf{Metric} & \textbf{Wilcoxon} & \textbf{p} & \textbf{Effect Size $r$} \\
\midrule
Relevance  & 32.0 & 0.08640 & 0.465 \\
Enjoyment  & 7.0  & \textbf{0.01617} & 0.789 \\
Difficulty & 35.0 & 0.57273 & 0.427 \\
\bottomrule
\end{tabular}
\caption{Wilcoxon signed-rank test results comparing Emotional Arc to Baseline for each user metric (one-sided, hypothesis: Emotional Arc \( > \) Baseline). Statistically significant $p$-values ($p < 0.05$) are shown in bold.}
\label{tab:wilcoxon}
\end{table}

\begin{figure*}[t]
\centering
\includegraphics[width=\textwidth]{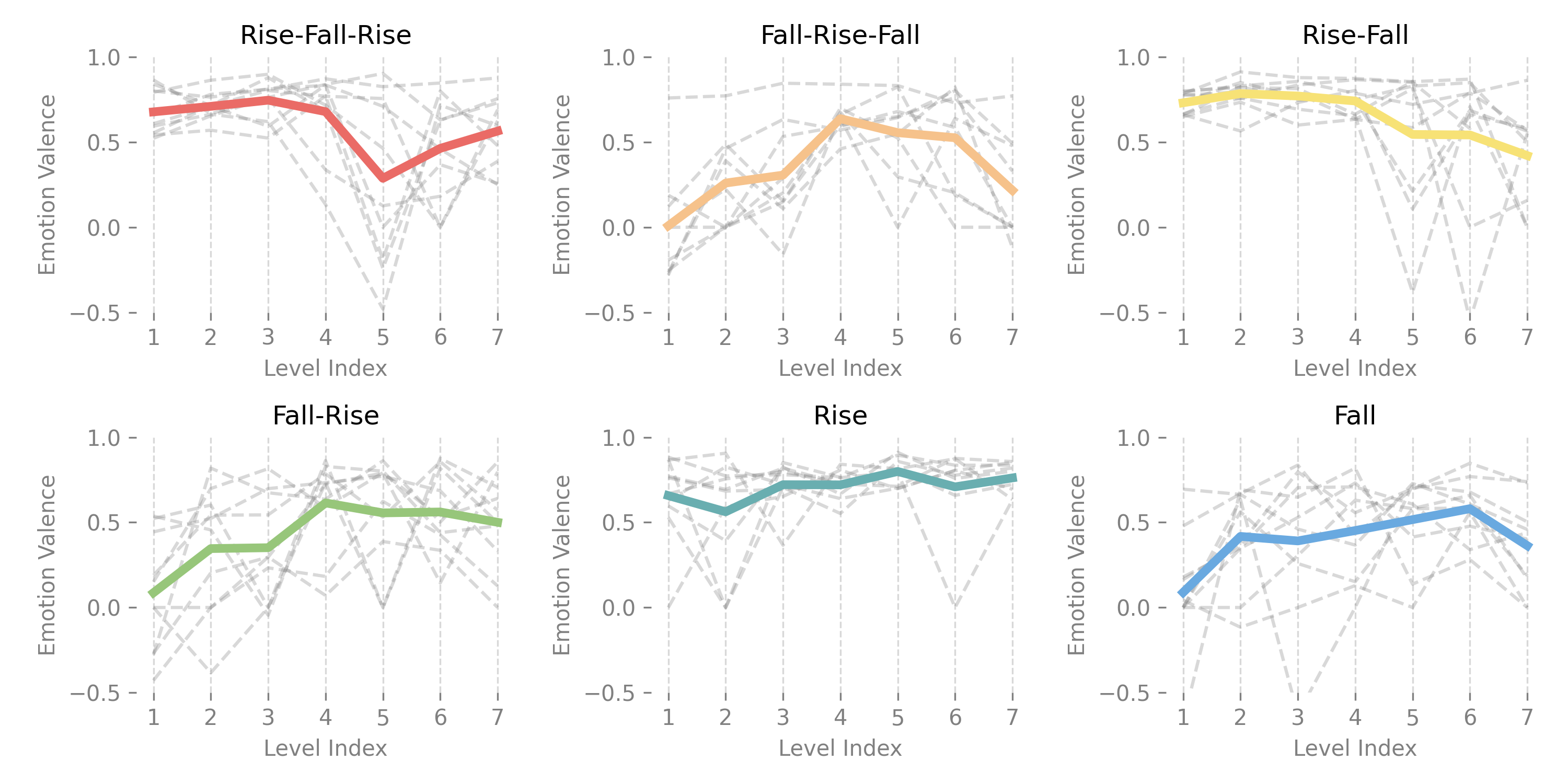}
\caption{
Emotion valence trajectories for six canonical emotional arcs across story levels (1–7). Each subplot corresponds to a different arc type, with gray dashed lines representing individual generated stories and colored bold lines showing the mean valence trajectory. Valence scores were computed using a multi-label emotion classifier and aggregated by story node. The results reveal distinctive curve shapes for each arc type, validating the alignment between intended narrative structure and generated emotional flow.
}
\label{fig:sentiment}
\end{figure*}

\section{Results}

\paragraph{User Ratings.} As shown in the boxplot (Figure \ref{fig:user_boxplot}), user ratings across the three evaluated attributes exhibit a wide spread, indicating substantial variability in perceived experiences. The presence of multiple outliers further suggests that some users rated the attributes in ways that deviated markedly from the central tendency.

The 95\% confidence intervals (CIs) reported in Table \ref{tab:bootstrap_fancy} provide additional insight into the reliability of user ratings across conditions. For the \textit{Enjoyment} metric, the intervals for the Baseline [3.500,4.938] and Emotional Arc [4.812,5.750] conditions do not overlap, offering strong evidence of a consistent and practically meaningful difference in user enjoyment. The substantial overlap in the intervals for \textit{Relevance }and \textit{Difficulty} indicates greater variability in individual responses and less distinct separation between conditions.

Table \ref{tab:wilcoxon} summarizes the results of one-sided Wilcoxon signed-rank tests comparing the Emotional Arc to the Baseline condition. A statistically significant improvement was observed in \textit{Enjoyment} (p = 0.016), with a large effect size (r = 0.789), highlighting the robust positive impact of Emotional Arc design on user experience. For \textit{Relevance,} the p-value did not reach statistical significance (p = 0.086), although the effect size was moderate to large (r = 0.465). This finding aligns with the non-overlapping portions of the confidence intervals noted above. No significant difference was found for \textit{Difficulty} (p = 0.573), and the smaller effect size (r = 0.427) suggests minimal practical impact. Notably, none of the metrics showed negative effects, supporting the conclusion that the Emotional Arc condition improved or at least maintained user experience across all measured dimensions.

\paragraph{User Perception.} 13 out of 16 participants (81.25\%) correctly identified the emotional arc condition, indicating that arc structure was salient and perceivable without explicit cues.

\paragraph{Qualitative Feedback.} In addition to the quantitative results, interviews with 16 participants revealed consistent themes across narrative, gameplay, and system-level design.

\textit{Narrative Engagement.} Participants consistently reported that the Emotional Arc system provided a richer and more emotionally engaging narrative experience. Multiple users described the Arc stories as “more dramatic," “better written," and “more realistic." One user noted that Game 1 “added more details" and developed a story that was “adequate for a simple game," while another emphasized that Arc versions offered “branching and emotional description" that were lacking in the baseline. In contrast, the Baseline version was described by users as “rigid” or “linear and without much detail”, with less compelling character interactions.

\textit{Gameplay Design and Difficulty.} While narrative complexity improved under Arc conditions, participants noted shortcomings in gameplay clarity and mechanical feedback. Several users commented that Arc versions involved “too much text” and “unclear item usage,” and lacked actionable gameplay prompts. One user observed more explicit difficulty progression in Arc levels, where emotional arcs (e.g., \textit{Fall}) translated into higher challenge and narrative tension. Conversely, the Baseline version was praised for its “clearer instructions” and “practical interaction flow.” Notably, a user criticized the difficulty of an Arc boss fight, citing unfair advantages against the player, highlighting the need for improved difficulty balancing.

\textit{Generation Pipeline and Assets.} Participants expressed generally positive impressions of the underlying generation system. Two users highlighted the UI and asset quality, describing the character and item generation as “great" and appreciating the “one-click game generation" feel. However, asset censorship was flagged as an issue, suggesting a need for improved content filtering. The system’s ability to maintain narrative continuity (e.g., item carryover, character evolution) was also noted by some users as a strength, particularly in Arc versions.

\textit{User Preferences.} A majority of participants expressed an overall preference for the Arc-based experiences, citing more complex plots, richer character development, and enhanced emotional depth. Some users, however, preferred the Baseline’s simplicity, clarity, or shorter gameplay loops, indicating variability in user preferences depending on play style and narrative tolerance.

\paragraph{Sentiment Alignment.}
 Figure \ref{fig:sentiment} illustrates the emotion valence trajectories computed for the six canonical emotional arc types across seven narrative levels. Each subplot corresponds to one arc type, where gray dashed lines represent individual generated stories and the colored solid line indicates the average valence trajectory for that arc.

Across all non-monotonic arc types, the valence trajectories demonstrate clear structural alignment with their target emotional shapes, suggesting that the system is capable of producing emotionally coherent story progressions. In particular, arcs such as Rise–Fall–Rise and Fall–Rise–Fall exhibit greater curvature and higher inter-story variance, as reflected by the wider spread of the dashed lines. This pattern reflects the inherent complexity and variability of non-monotonic narrative dynamics.

In contrast, monotonic arc types (steady Rise, steady Fall) do not exhibit strong directional trends. Rather than showing a consistent upward or downward slope, the trajectories tend to remain relatively flat or fluctuate around a moderately positive baseline. If a strict interpretation of monotonicity is assumed, where each subsequent node is expected to show a consistently higher or lower valence than the previous one, the current results suggest that the model does not reliably enforce such directional constraints. This limitation may stem from its inability to enforce step-wise emotional progression in the absence of explicit valence supervision.

The overall valence levels are consistently skewed toward the positive range across all arc types. Most individual points fall above zero, suggesting a systemic bias in either the story generation or the emotion classifier toward positive affect. This general positivity may obscure finer-grained emotional distinctions, particularly in arcs intended to evoke negative or downward trajectories.

\section{Discussion}
\subsection{RQ1: Is the Emotional Arc Design Effective?}
Our findings indicate that the Emotional Arc framework meaningfully enhances narrative engagement. Both statistical and subjective measures support that user ratings of \textit{Enjoyment} were significantly higher under the Emotional Arc condition, suggesting that aligning gameplay progression with canonical emotional arcs can increase the experiential richness of narrative-driven games.

In contrast, differences in \textit{Relevance} and \textit{Difficulty} did not reach statistical significance. One likely explanation for the \textit{Relevance} outcome lies in the diversity of player archetypes. Narrative-focused players are more attuned to story coherence and emotional tone, aligning with the arc-based design. In contrast, goal-oriented or action-centric players often prioritize mechanical engagement, assessing \textit{Relevance} through gameplay variables such as enemy frequency, combat flow, or spatial structure, the elements not directly optimized by emotional shaping.

The lack of a significant difference in \textit{Difficulty} may be attributed to limitations in the system’s implementation. Baseline levels did not include consistent difficulty modulation, which reduced their suitability as a comparison for the Emotional Arc condition. In addition, confounding factors such as the presence of ranged-attack enemies in both conditions may have overshadowed more subtle shifts in difficulty, especially given the relatively simple mechanics of the prototype.

Taken together, these findings affirm the efficacy of emotional arcs for enhancing story perception while also highlighting the importance of designing complementary gameplay systems to fully realize their potential.

\subsection{RQ2: Are Emotional Arcs Perceivable to Humans and Machines?}
The majority of participants (81.25\%) accurately identified the arc-based condition without explicit instruction, indicating that emotional trajectories were sufficiently salient to influence user perception when embedded within procedurally generated game levels.

However, perceptibility varied by presentation and player preference. Several participants noted that Arc-generated narratives could become overly verbose, with metaphorical or descriptive language interrupting the gameplay rhythm. Others preferred the Baseline version’s conciseness, which allowed for smoother progression and clearer focus on interaction. This trade-off illustrates that the visibility of emotional arcs is not binary, but contingent on textual pacing, delivery format, and the player’s interpretive frame.

The post-hoc emotion valence trajectories showed structural patterns that closely mirrored the canonical shapes of the six emotional arcs, especially for non-monotonic forms such as Rise–Fall–Rise and Fall–Rise–Fall. These trajectories, derived from an independent emotion classification model, provide external validation that the generated stories preserved the intended emotional dynamics.

The convergence between human perception and machine-derived emotional structure point toward a promising direction for future integration: embedding emotional arc cues not only in text but also in gameplay events, environmental triggers, and audiovisual dynamics. Such multimodal reinforcement may improve perceptibility while reducing narrative friction.

\subsection{System Integration and Gameplay Trade-offs}
Our current system was deployed within a single ARPG template, selected for its alignment with node-based storytelling and interactive level design. While this format supports story-level emotional arcs, several limitations emerged. Some participants expressed that gameplay interactions lacked depth or consequence, weakening the link between narrative content and player agency. Others perceived a disconnect between emotionally complex plots and comparatively simple mechanics, reducing immersion.

These limitations emphasize the need for tighter coupling between narrative structure and gameplay expression. Emotional arcs should ideally manifest not only through text but also through systemic behaviors such as adaptive enemy behavior, context-sensitive events, and branching consequences. The present pipeline uses static difficulty values derived from LLM prompts; future iterations could benefit from dynamic difficulty adjustment systems that align gameplay pacing with emotional arc trajectories in real time.

\section{Limitations}
This work has several limitations that warrant consideration. The user study involved only 16 participants with relatively homogeneous backgrounds, which limits statistical power and restricts the ability to analyze the relationship between emotional arcs and factors such as occupation, gender, or prior gaming experience. The diversity of player archetypes and individual differences in narrative sensitivity were not explicitly modeled or controlled, potentially affecting the generalizability of the results.

On the system side, the Baseline condition did not include consistent difficulty modulation, and the presence of ranged-attack enemies in both conditions may have confounded comparative analysis of difficulty. Additionally, the current implementation focuses on narrative shaping primarily through text, without incorporating systemic gameplay responses such as adaptive enemy behavior, context-sensitive events, or branching consequences. This limits the depth of player agency and the alignment between narrative and mechanical experiences.

Finally, the system was evaluated solely within an ARPG template, which may not generalize to other game genres or broader interactive entertainment contexts. Further research is needed to validate the applicability of emotional arc–guided generation in different gameplay formats and design paradigms.

\section{Future Work}
Building on these limitations, we identify several directions for future research. First, the generalizability of our findings is limited by genre: ARPGs offer a favorable balance between story and gameplay, but may not reflect the dynamics of other genres. Extending this framework to visual novels, simulation games, or roguelikes could reveal new affordances and constraints.

Second, integrating non-verbal emotional signals, such as audio, visuals, and pacing, may provide a richer substrate for communicating emotional progression. 

Third, personalized narrative scaffolding through player modeling and adaptive feedback could help align story structure with individual user preferences, paving the way for emotionally adaptive interactive storytelling.

\section{Conclusion}
This work presents a novel framework for emotional arc-guided procedural game level generation, grounded in narratological theory and operationalized through large language models. By embedding emotional trajectories into both narrative and gameplay elements, our system produces story graphs and interactive experiences that align with canonical rise–fall structures. Empirical evaluation demonstrates that emotional arc integration significantly enhances user enjoyment and narrative engagement, while remaining perceivable to both players and machine-based sentiment models. At the same time, our findings underscore key challenges in gameplay–narrative alignment, difficulty modulation, and player diversity. As generative AI continues to reshape the design space of interactive storytelling, emotional arcs offer a promising yet underexplored scaffold for ensuring coherence, immersion, and emotional resonance. Future work will further investigate multimodal expression, adaptive difficulty, and cross-genre applicability to broaden the impact of emotionally structured narrative generation in games.



\bibliography{aaai24}

\end{document}